%% file: propnet.tex
\documentclass[letterpaper, 10 pt, conference]{ieeeconf}  

\IEEEoverridecommandlockouts                              

\input{packages}
\input{macros}

\title{\LARGE \bf
Propagation Networks for Model-Based Control \\Under Partial Observation
}

\author{Yunzhu Li, Jiajun Wu, Jun-Yan Zhu, Joshua B. Tenenbaum, Antonio Torralba, and Russ Tedrake
\thanks{Y. Li, J. Wu, J.-Y. Zhu, J. B. Tenenbaum, A. Torralba, and R. Tedrake are with the Computer Science and Artificial Intelligence Laboratory (CSAIL) at Massachusetts Institute of Technology, Cambridge, MA, USA}
}

\begin{document}

\maketitle
\thispagestyle{empty}
\pagestyle{empty}

\input{text/abstract.tex}

\input{text/intro.tex}
\input{text/related_work.tex}

\input{text/dynamics.tex}
\input{text/control.tex}
\input{text/experiments.tex}

\input{text/conclusion.tex}
\input{text/ack.tex}

\bibliographystyle{IEEEtran}
\bibliography{propnet}

\end{document}

%% file: packages.tex
\usepackage{color,xcolor}
\usepackage{epsfig}
\usepackage{graphicx}

\usepackage{adjustbox}
\usepackage{array}
\usepackage{booktabs}
\usepackage{colortbl}
\usepackage{float,wrapfig}
\usepackage{hhline}
\usepackage{multirow}
\usepackage{subcaption} 
\usepackage[font={footnotesize}]{caption}

\usepackage{amsmath,amsfonts,amssymb}
\usepackage{bm}
\usepackage{microtype}

\usepackage{changepage}
\usepackage{extramarks}
\usepackage{fancyhdr}
\usepackage{lastpage}
\usepackage{setspace}
\usepackage{soul}
\usepackage{xspace}

\usepackage{url}

\usepackage{algorithm, algorithmic}
\usepackage{enumerate}
\usepackage{todonotes} 

%% file: macros.tex
\newcolumntype{L}[1]{>{\raggedright\let\newline\\\arraybackslash\hspace{0pt}}m{#1}}
\newcolumntype{C}[1]{>{\centering\let\newline\\\arraybackslash\hspace{0pt}}m{#1}}
\newcolumntype{R}[1]{>{\raggedleft\let\newline\\\arraybackslash\hspace{0pt}}m{#1}}

\newcommand{\sect}[1]{Section~\ref{#1}}

\newcommand{\fig}[1]{Fig.~\ref{#1}}

\newcommand{\ignore}[1]{}

\makeatletter
\DeclareRobustCommand\onedot{\futurelet\@let@token\@onedot}
\def\@onedot{\ifx\@let@token.\else.\null\fi\xspace}

\def\eg{e.g\onedot} 
\def\ie{i.e\onedot}

\def\etal{et al\onedot}
\makeatother

\definecolor{MyDarkBlue}{rgb}{0,0.08,1}
\definecolor{MyDarkGreen}{rgb}{0.02,0.6,0.02}
\definecolor{MyDarkRed}{rgb}{0.8,0.02,0.02}
\definecolor{MyDarkOrange}{rgb}{0.40,0.2,0.02}
\definecolor{MyPurple}{RGB}{111,0,255}
\definecolor{MyRed}{rgb}{1.0,0.0,0.0}
\definecolor{MyGold}{rgb}{0.75,0.6,0.12}
\definecolor{MyDarkgray}{rgb}{0.66, 0.66, 0.66}

\newcommand{\myparagraph}[1]{\vspace{0pt}\paragraph{#1}}

\newcommand{\Model}{Propagation Networks\xspace}
\newcommand{\model}{propagation networks\xspace}
\newcommand{\modelshort}{PropNet\xspace}
\newcommand{\modelshorts}{PropNets\xspace}
\newcommand{\basemodelshort}{Vanilla PropNet\xspace}

%% file: text/abstract.tex
\begin{abstract}
    There has been an increasing interest in learning dynamics simulators for model-based control. 
    Compared with off-the-shelf physics engines, a learnable simulator can quickly adapt to unseen objects, scenes, and tasks. 
    However, existing models like interaction networks only work for fully observable systems; they also only consider pairwise interactions within a single time step, both restricting their use in practical systems. We introduce \Model (\modelshort), a differentiable, learnable dynamics model that handles partially observable scenarios and enables instantaneous propagation of signals beyond pairwise interactions. 
    Experiments show that our \model not only outperform current learnable physics engines in forward simulation, but also achieve superior performance on various control tasks. Compared with existing model-free deep reinforcement learning algorithms, model-based control with \model is more accurate, efficient, and generalizable to new, partially observable scenes and tasks.
\end{abstract}

%% file: text/intro.tex
\section{Introduction}

Physics engines are critical for planning and control in robotics. To plan for a task, a robot may use a physics engine to simulate the effects of different actions on the environment and then select a sequence of actions to reach a desired goal configuration. The utility of the resulting action sequence depends on the accurate prediction of the physics engine; so a high-fidelity physics engine plays a critical role in robot planning. Most physics engines used in robotics, such as Mujoco~\cite{Todorov2012MuJoCo}, Bullet~\cite{Coumans2010Bullet}, and Drake~\cite{drake}, use approximate contact models, and recent studies \cite{Kolbert2016}, \cite{Yu2016More}, \cite{Fazeli2017Fundamental} have demonstrated discrepancies between their predictions and real-world data. These mismatches prevent the above physics engines from solving contact-rich tasks.

Recently, researchers have started building general-purpose neural physics simulators, aiming to approximate complex physical interactions with neural networks~\cite{Battaglia2016Interaction,Chang2017compositional}. 
They have succeeded to model the dynamics of both rigid bodies and deformable objects (\eg, ropes). More recent work has used interaction networks for discrete and continuous control~\cite{Racaniere2017Imagination,Hamrick2017Metacontrol,Pascanu2017Learning,sanchez2018graph}.

Interaction networks, however, have two major limitations. First, interaction nets only consider pairwise interactions between objects, restricting its use in real-world scenarios, where simultaneous multi-body interactions often occur. Typical examples include Newton's cradle (\fig{fig:teaser_a}) or rope manipulation (\fig{fig:teaser_b}). Second, they need to observe the full states of a environment; however, many real-world control tasks involve dealing with partial observable states. \fig{fig:teaser_c} shows an example, where a robot wants to push a set of blocks into a target configuration; however, only the red blocks in the top layer are visible to the camera.

In this paper, we introduce \Model (\modelshort), a differentiable, learnable engine that simulates multi-body object interactions. \modelshort handles partially observable situations by operating on a latent dynamics representation; it also enables instantaneous propagation of signals beyond pairwise interactions using multi-step effect propagation. Specifically, by representing a scene as a graph, where objects are the vertices and object interactions are the directed edges, we initialize and propagate the signals through the directed paths in the interaction graph at each time step.

Experiments demonstrate that \modelshort consistently outperforms interaction networks in forward simulation. \modelshort's ability to accurately handle partially observable states brings significant benefits for control. Compared with interaction nets and state-of-the-art model-free deep reinforcement learning algorithms, model-based control using \model is more sample-efficient, accurate, and generalizes better to new, partially observable scenarios.\footnotetext{Our project page: \url{http://propnet.csail.mit.edu}}

\input{figText/teaser.tex}

%% file: figText/teaser.tex
\begin{figure}[t]
  \centering
  \begin{subfigure}[t]{.45\linewidth}
    \centering
    \includegraphics[width=.8\textwidth]{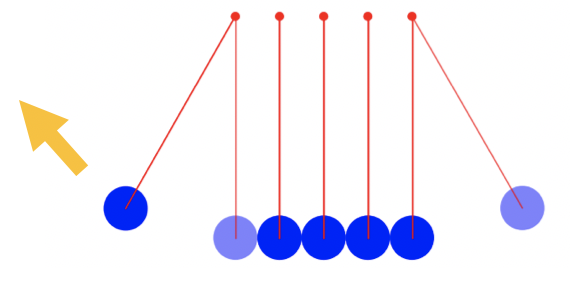}
    \caption{\label{fig:teaser_a} \footnotesize Newton's Cradle}
  \end{subfigure}
  \begin{subfigure}[t]{.45\linewidth}
    \centering
    \includegraphics[width=.55\textwidth]{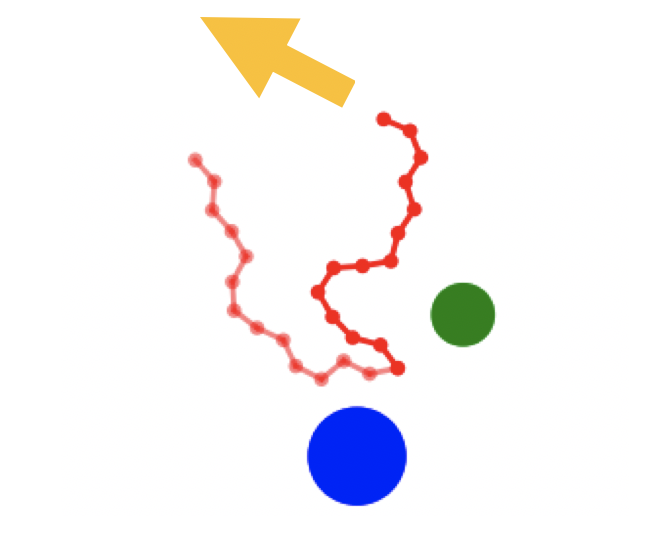}
    \caption{\label{fig:teaser_b} \footnotesize Rope Manipulation}
  \end{subfigure}
  \begin{subfigure}{\linewidth}
    \centering
    \includegraphics[width=.65\textwidth]{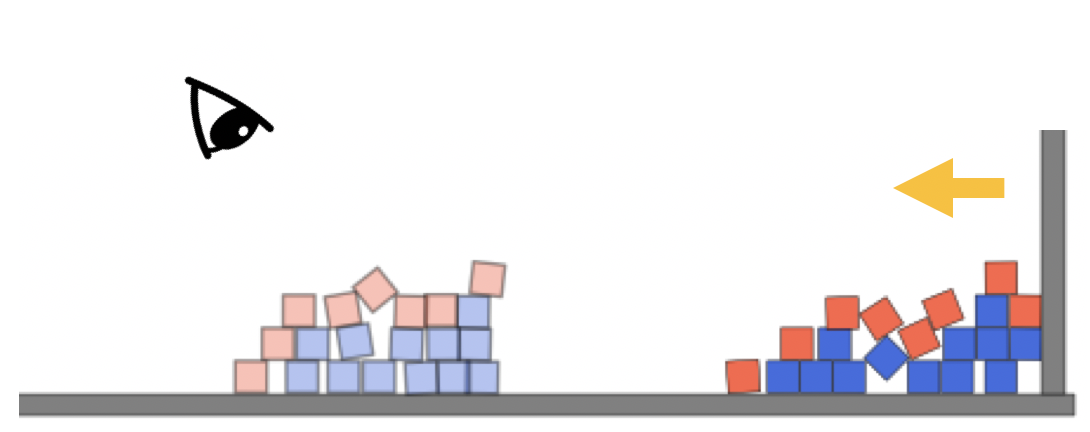}
    \caption{\label{fig:teaser_c} \footnotesize Box Pushing}
  \end{subfigure}
  \caption{{\bf Challenges for existing differentiable physics simulators:} Modeling the dynamics of (a) Newton's cradle or (b) a rope requires instantaneous propagation of multi-object interaction. For (a), our goal is to control the leftmost ball so that rightmost ball hits the target (transparent). For (b), our goal is to control the rope to reach the target (transparent), while the blue and green circles are fixed obstacles. (c) Pushing a group of boxes to a target configuration requires dynamics modeling under partial observations. Here, the camera is looking down and only red blocks are observable.}
  \label{fig:teaser}
\end{figure}

%% file: text/related_work.tex
\section{Related Work}

\input{figText/cradle.tex}

\subsection{Differentiable Physics Simulators} 

In recent years, researchers have been building differentiable physics simulators in various forms~\cite{Todorov2012MuJoCo,drake,Ehrhardt2017Taking,degrave2016differentiable}. For example, approximate, analytical differentiable rigid body simulators~\cite{degrave2016differentiable,de2018modular} have been deployed for tool manipulation and tool-use planning~\cite{toussaint2018differentiable}. 

Among them, two notable efforts on learning differentiable simulators include interaction networks~\cite{Battaglia2016Interaction} and neural physics engines~\cite{Chang2017compositional}. These methods restrict themselves to pairwise interactions for generalizability. However, this simplification limits their ability to handle simultaneous, multi-body interactions. In this work, we tackle this problem by learning to propagate the signals multiple steps on the interaction graph. Gilmer~\etal~\cite{gilmer2017neural} have recently explored message passing networks, but with a focus on quantum chemistry. 

\subsection{Model-Predictive Control with a Learned Simulator}

Recent work on model-predictive control with deep networks~\cite{Lenz2015DeepMPC,Gu2016Continuous,nagabandi2017neural,farquhar2017treeqn,srinivas2018universal} often learns an abstract-state transition function, instead of an explicit account of environments~\cite{Silver2017predictron,Oh2017Value}. Subsequently, they use the learned model or value function to guide the training of the policy network. Instead, \modelshort learns a general physics simulator that takes raw object observations (\eg, positions, velocities) as input. We then integrate it into classic trajectory optimization algorithms for control. 

A few recent papers exploit the power of interaction networks for planning and control. Many of them use interaction networks to \emph{imagine}---rolling out approximate predictions---to facilitate training a policy network~\cite{Racaniere2017Imagination,Hamrick2017Metacontrol,Pascanu2017Learning}. In contrast, we use propagation networks as a learned dynamics simulator and directly optimize trajectories for continuous control. By separating model learning and control, our model generalizes better to novel scenarios. Recently, Sanchez-Gonzalez~\etal~\cite{sanchez2018graph} also explored applying interaction networks for control. Compared with them, our \model can handle simultaneous multi-body interactions and deal with partially observable scenarios.

%% file: figText/cradle.tex
\begin{figure*}[t]
    \centering
    \includegraphics[width=.8\linewidth]{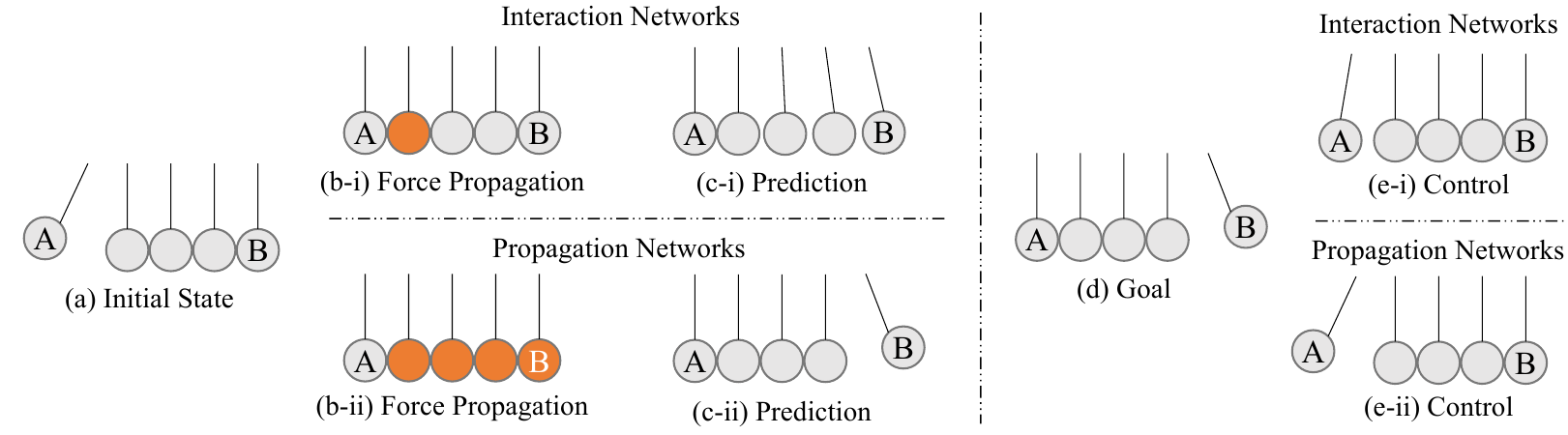}
    \caption{{\bf Newton's Cradle.} (a) shows the initial states of a Newton's cradle, based on which both the Interaction Networks and Propagation Networks try to predict future states; (b-i) The Interaction Networks can only propagate the force along a single relation at a time step, thus results in a false prediction (c-i); (b-ii) Our proposed method can propagate the force correctly which leads to the correct prediction (c-ii); (d) A downstream task where we aim to achieve a specific goal using the learned model; (e-i) Model-based control methods fail to produce the correct control using Interaction Networks while (e-ii) our model can provide the desired control signal.}
    \label{fig:cradle}
\end{figure*}

%% file: text/dynamics.tex
\section{Learning the Dynamics}

\subsection{Preliminaries}

We assume that the interactions within a physical system can be represented as a directed graph, $G=\langle O, R\rangle$, where vertices $O$ represent the objects, and edges $R$ correspond to their relations (\fig{fig:pn}). Graph $G$ can be represented as
\begin{equation}
    O = \{o_i\}_{i=1\dots |O|} \quad \quad \quad R = \{r_k\}_{k=1\dots |R|}
\end{equation}
Specifically, $o_i = \langle x_i, a^o_i, p_i \rangle $, where $x_i = \langle q_i, \dot{q}_i \rangle$ is the state of object $i$, containing its position $q_i$ and velocity $\dot{q}_i$. $a^o_i$ denote its attributes (\eg, mass, radius), and $p_i$ is the external force on object $i$. For the relations, we have
\begin{equation}
    r_k = \langle u_k, v_k, a^r_k \rangle, \quad 1 \leq u_k, v_k \leq |O|,
\end{equation}
where $u_k$ is the receiver, $v_k$ is the sender, and $a^r_k$ is the type and  attributes of relation $k$ (\eg, collision, spring connection).

Our goal is to build a learnable physics engine to approximate the underlying physical interactions. We can then use it to infer the system dynamics and predict the future from the observed interaction graph $G$:
\begin{equation}
\label{eq:dynamics}
    G_{t+1}=\phi(G_t),
\end{equation}
where $G_t$ denotes the scene states at time $t$. We aim to learn $\phi(\cdot)$, a learnable dynamics model, to minimize $\|G_{t+1} - \phi(G_t)\|_2$.

Below we review our baseline model Interaction Networks (IN)~\cite{Battaglia2016Interaction}. IN is a general-purpose, learnable physics engine, performing object- and relation-centric reasoning about physics. IN defines an object function $f_O$ and a relation function $f_R$ to model objects and their relations in a compositional way. The future state at time $t + 1$ is predicted as
\begin{align}
\label{eq:in}
    e_{k, t} & = f_R(o_{u_k, t}, o_{v_k, t}, a^r_k), \quad k = 1 \dots |R|, \nonumber\\
    \hat{o}_{i, t+1} & = f_O(o_{i, t}, \sum_{k \in \mathcal{N}_i}e_{k, t}), \quad i = 1\dots |O|,
\end{align}
where $o_{i, t} = \langle x_{i, t}, a^o_i, p_{i, t} \rangle$ denotes object $i$ at time $t$, $u_k$ and $v_k$ are the receiver and sender of relation $r_k$, and $\mathcal{N}_i$ denotes the relations where object $i$ is the receiver.

\subsection{\Model}

IN defines a flexible and efficient model for explicit reasoning of objects and their relations in a complex system. It can handle a variable number of objects and relations and has performed well in domains like n-body systems, bouncing balls, and falling strings. However, one fundamental limitation of IN is that at every time step $t$, it only considers local information in the graph $G$ and cannot handle instantaneous propagation of forces, such as Newton's cradle shown in \fig{fig:cradle}, where ball A's impact produces a compression wave that propagates through the balls immediately~\cite{stewart2000rigid}. As force propagation is a common phenomenon in rigid-body dynamics, this shortcoming has limited IN's practical applicability.

To address the above issues, we propose \Model (\modelshort) to handle the instantaneous propagation of forces efficiently.  
Our method is inspired by message passing, a classic algorithm in graphical models~\cite{pearl2014probabilistic}.

\subsubsection{Effect propagation}

Effect propagation requires multi-step message passing along the directed edges in graph $G$. Forces ejected from ball A (\fig{fig:cradle}) should be propagated through the connected balls to ball B within a single time step. 
Force propagation is hard to analyze analytically for complex scenes. Therefore, we let \modelshort learn to decide whether an effect should be propagated further or withheld.

At time $t$, we denote the propagating effect from relation $k$ at propagation step $l$ as $e^l_{k, t}$, and the propagating effect from object $i$ as $h^l_{i, t}$. Here, we have $1 \leq l \leq L$, where $L$ is the maximum propagation steps within each step of the simulation. Propagation can be described as 

\begin{align}
\label{eq:pn-v1}
    & \text{Step 0: } && h_{i,t}^0 = \mathbf{0}, \quad i = 1\dots |O|,\nonumber \\
    & \text{Step $l = 1,\dots,L$: } && e^{l}_{k, t} = f^{l}_R(o_{u_k, t}, o_{v_k, t}, a^r_k, h^{l-1}_{u_k, t}, h^{l-1}_{v_k, t}), \nonumber \\
    &&& k = 1 \dots |R|,  \nonumber \\
    &&& h^{l}_{i, t} = f^{l}_O(o_{i, t}, \sum_{k \in \mathcal{N}_i}e^{l}_{k, t}), \nonumber \\
    &&& i = 1\dots |O|,\nonumber \\
    & \text{Output: } && \hat{o}_{i, t+1} = h^L_{i, t}, \quad  i = 1\dots |O|,
\end{align}

\noindent
where $f^l_O(\cdot)$ denotes the object propagator at propagation step $l$， and $f^l_R(\cdot)$ denotes the relation propagator. Depending on the complexity of the task, the network weights can be shared among propagators at different propagation steps.

We name this model \basemodelshort. Experimental results show that the selection of $L$ is task-specific, and usually a small $L$ (\eg, $L=3$) can achieve a good trade-off between the performance and efficiency.

\subsubsection{Object- and relation-encoding with residual connections}
\label{sec:encoding_and_residual}

We notice that \basemodelshort is not efficient for fast online control. As information such as states $o_{i, t}$ and attributes $a^r_k$ are fixed at a specific time step, they can be shared without re-computation between each sequential propagation step. Hence, inspired by the ideas on fast RNNs training~\cite{lei2017training,bradbury2016quasi}, we propose to encode the shared information beforehand and reuse them along the propagation steps. We denote the encoder for objects as $f^{\text{enc}}_O(\cdot)$ and the encoder for relations as $f^{\text{enc}}_R(\cdot)$. Then,
\begin{equation}
\label{eq:encode}
    c^o_{i, t} = f^{\text{enc}}_O(o_{i, t}), \quad \quad c^r_{k, t} = f^{\text{enc}}_R(o_{u_k, t}, o_{v_k, t}, a^r_k).
\end{equation}
In practice, we add residual links~\cite{he2016deep} between adjacent propagation steps that connect $h^l_{i, t}$ and $h^{l-1}_{i, t}$. This helps address gradient vanishing and exploding problem, and provides access to historical effects. The update rules become
\begin{equation}
\label{eq:pn}
\begin{split}
    e^l_{k, t} &= f^l_R(c^r_{k, t}, h^{l-1}_{u_k, t}, h^{l-1}_{v_k, t}), \\
    h^l_{i, t} &= f^l_O(c^o_{i, t}, \sum_{k \in \mathcal{N}_i}e^{l}_{k, t}, h^{l-1}_{i, t}),
\end{split}
\end{equation}
where propagators $f^l_O(\cdot)$ and $f^l_R(\cdot)$ now take a new sets of inputs, which is different from \basemodelshort.

Based on the assumption that the effects between propagation steps can be represented as simple transformations (\eg, identity-mapping in Newton's cradle), we can use small networks as function approximators for the propagators $f^l_O(\cdot)$ and $f^l_R(\cdot)$ for better efficiency.
We name this updated model \Model (\modelshort).

\input{figText/pn.tex}

\input{figText/pn_partial.tex}

\subsection{Partially Observable Scenarios}

For many real-world situations, however, it is often hard or impossible to estimate the full state of environments. We extend Eqn.~\ref{eq:dynamics} using \modelshorts to handle such partially observable cases by operating on a latent dynamics model:
\begin{equation}
\label{eq:dynamics_latent}
    \tau(G_{t+1})=\phi(\tau(G_t)),
\end{equation}
where $\tau(\cdot)$ is an encoding function that maps the current observation to a latent representation.
As shown in Figure~\ref{fig:pn_partial}b, $\tau(\cdot)$ consists of two parts: first, \modelshorts $g(\cdot)$ that map the current observation to object-centric representations; second, $\omega(\cdot)$ that aggregates the object-centric representations into a fixed-dimensional global representation. We use a global representation for partially observable cases, because the number and set of observable objects vary over time, making it hard to define object-centric dynamics.
In fully observable environments, $\tau(\cdot)$ reduces to an identity mapping and the dynamics is defined on the object level over the state space (Eqn.~\ref{eq:dynamics} and Fig.~\ref{fig:pn_partial}a). To train such a latent dynamics model, we seek to minimize the loss function: $\mathcal{L}_{\text{forward}} = \|\tau(G_{t+1}) - \phi(\tau(G_t))\|_2$.

In practice, we use a small history window of length $T_\text{history}$ for the state representation, \ie, the input to $\phi(\cdot)$ is the concatenation of $\tau(G_t), \tau(G_{t-1}), ..., \tau(G_{t-T_\text{history}+1})$.

Using the above loss alone leads to trivial solutions such as $\phi(x)=\tau(x)=0$ for any valid $x$. We tackle this based on an intuitive idea: an ideal encoding function $\tau(\cdot)$ should be able to reserve information about the scene observation. Hence, we use an aggregation function $\omega(\cdot)$ that has no learnable parameters like summation or average and introduce a decoding function $\psi(\cdot)$ to ensure a nontrivial $\tau(\cdot)$ by minimizing an additional auto-encoder reconstruction loss~\cite{hinton2006reducing}: $\mathcal{L}_{\text{encode}} = \|G - \psi(g(G))\|_2$, where $\psi(\cdot)$ is realized as \modelshorts. The full model is shown in Figure~\ref{fig:pn_partial}b.

%% file: figText/pn.tex
\begin{figure}[t]
    \centering
    \includegraphics[width=.9\linewidth]{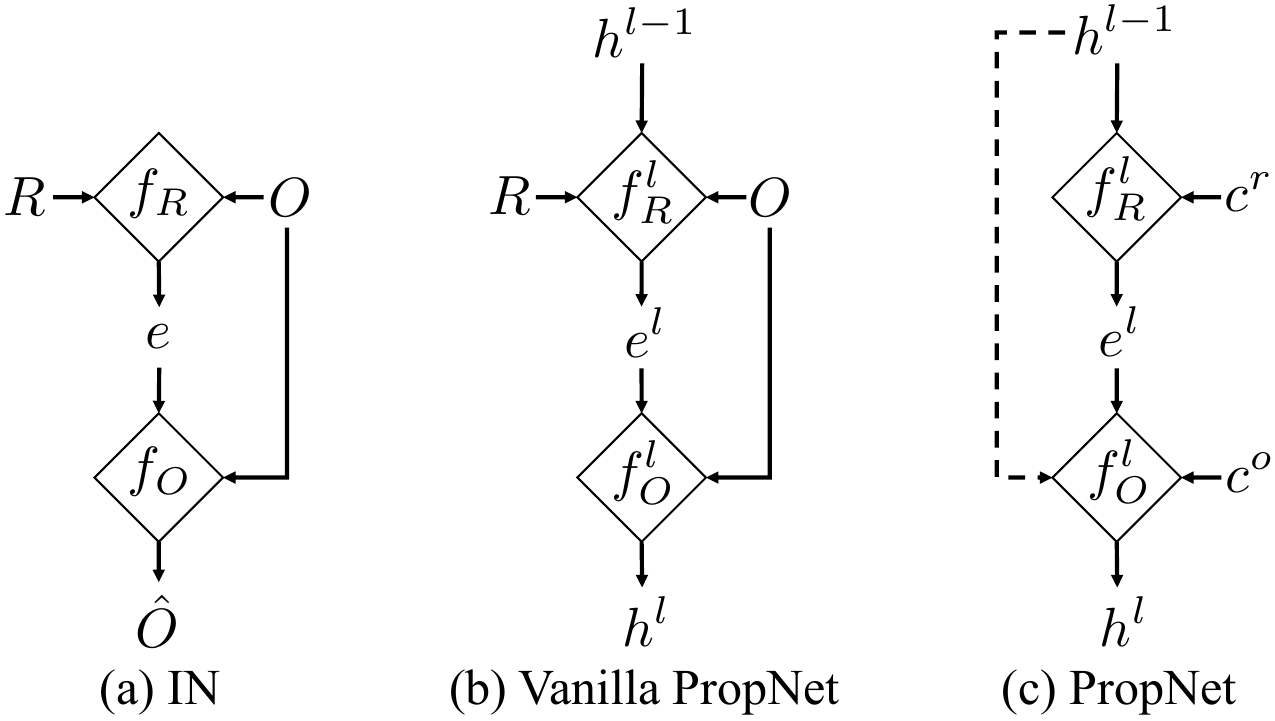}
    \caption{{\bf Graphical illustration of the models.} (a) The structure of Interaction Networks as detailed in Eqn.~\ref{eq:in}; (b) The internal structure of \basemodelshort is described in Eqn.~\ref{eq:pn-v1}, where the effects $e^l$ and $h^l$ are propagated through the propagators $f^l_O$ and $f^l_R$ along the directed relations in the graph $G$; (c) The shared object encoding $c^o$ and relation encoding $c^r$ are inputs to the internal modules, where there are also residual connections for better effect propagation as described in Eqn.~\ref{eq:encode} and~\ref{eq:pn}.}
    \label{fig:pn}
\end{figure}

%% file: figText/pn_partial.tex
\begin{figure}[t]
    \centering
    \includegraphics[width=.9\linewidth]{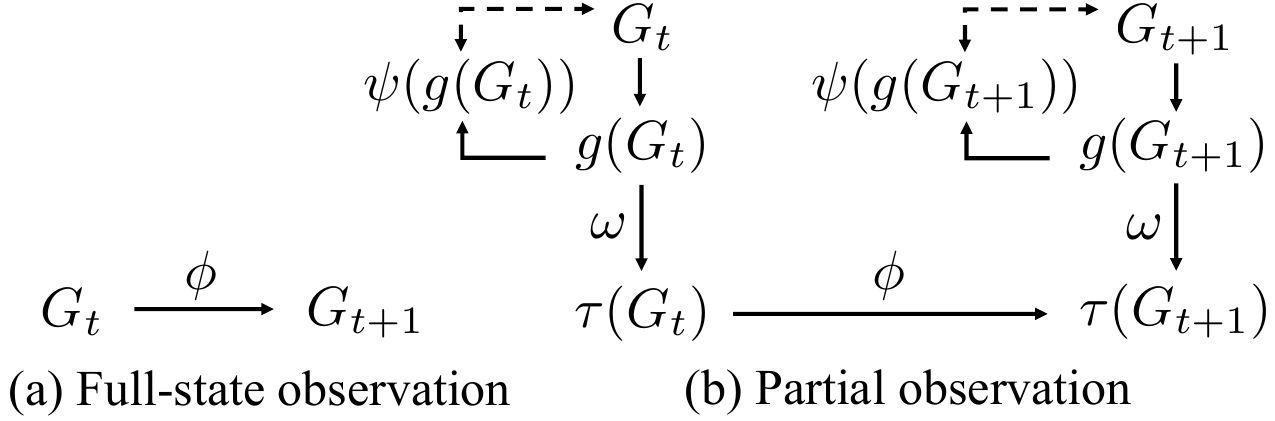}
    \caption{{\bf Comparison between fully- and partially-observable scenarios.}  (a) Forward model for fully observable environments (Eqn.~\ref{eq:dynamics}). (b) For partially observable scenarios, we first map the observation to a latent space using function $\tau(\cdot)$, and then specify the forward dynamics over the latent space using $\phi(\cdot)$ as described in Eqn.~\ref{eq:dynamics_latent}. $\tau(\cdot)$ consists of $g(\cdot)$ and $\omega(\cdot)$, where $g(\cdot)$ maps the observation to object-based representations, which are then aggregated to a global representation using $\omega(\cdot)$. A decoding function $\psi(\cdot)$ maps the encoding back to the original observation space to ensure a nontrivial encoding.}
    \vspace{-10pt}
    \label{fig:pn_partial}
\end{figure}

%% file: text/control.tex
\section{Control Using Learned Dynamics}

Compared to model-free approaches, model-based methods offer many advantages, such as generalization and sample efficiency, as it can approximate the policy gradient or value estimation without exhausted trials and errors.

However, an accurate model of the environment is often hard to specify and brings significant computational costs for even a single-step forward simulation. It would be desirable to learn to approximate the underlying dynamics from data.

A learned dynamics model is naturally differentiable. Given the model and a desired goal, we can perform forward simulation, optimizing the control inputs by minimizing a loss between simulated results and a goal. The model can also estimate the uncertain attributes online by minimizing the difference between predicted future and actual outcome. Alg.~\ref{alg:control} outlines our control algorithm, which provides a natural testbed for evaluating the dynamics models.

\myparagraph{Model predictive control using shooting methods}

Let $\mathcal{G}_g$ be our goal and $\hat{u}_{1:T}$ be the control inputs (decision variables), where $T$ is the time horizon. These task-specific control inputs are part of the dynamics graph. Typical choices include observable objects' initial velocity/position and external forces/attributes on objects/relations. We denote the graph encoding as $G^\tau = \tau(G)$, and the resulting trajectory after applying the control inputs as $\mathcal{G} = \{G_i^\tau\}_{i=1:T}$. The task here is to determine the control inputs by minimizing the gap between the actual outcome and the specified goal $\mathcal{L}_{\text{goal}}(\mathcal{G}, \mathcal{G}_g)$.

Our \model can do forward simulation by taking the dynamics graph at time $t$ as input, and produce the graph at next time step, $\hat{G}_{t+1}^\tau = \phi(G_t^\tau)$. Let's denote the forward simulation from time step $t$ as $\hat{\mathcal{G}} = \{ \hat{G}_i^\tau\}_{i=t+1\dots T}$ and the history until time $t$ as $\bar{\mathcal{G}} = \{G_i^\tau\}_{i=1\dots t}$. We can back-propagate from the loss $\mathcal{L}_g(\bar{\mathcal{G}} \cup \hat{\mathcal{G}}, \mathcal{G}_g)$ and use stochastic gradient descent (SGD) to update the control inputs. This is known as the shooting method in trajectory optimization~\cite{tedrakeunderactuated}.

If the time horizon $T$ is too long, the learned model might deviate from the ground truth due to accumulated prediction errors. Hence, we use Model-Predictive Control (MPC)~\cite{camacho2013model} to stabilize the trajectory by doing forward simulation at every time step as a way to compensate the simulation error.

\myparagraph{Online adaptation}

In many situations, inherent attributes such as masses, friction, and damping are not directly observable. Instead, we can interact with the objects and use \modelshort to estimate these attributes online (denoted as $A$) with SGD updates
by minimizing the difference between the predicted future states and the actual future states $\mathcal{L}_{\text{state}}(\hat{G}_{t}^\tau, G_{t}^\tau)$.
\begin{center}
\begin{algorithm}[t]
   \caption{Control on Learned Dynamics at Time Step $t$}
   \label{alg:control}
\begin{algorithmic}
   \STATE {\bfseries Input:} Learned forward dynamics model $\phi(\cdot)$ \\
   \quad Predicted dynamics graph encoding $\hat{G}_t^\tau$\\
   \quad Current dynamics graph encoding $G_t^\tau$\\
   \quad Goal $\mathcal{G}_g$, current estimation of the attributes $A$\\
   \quad Current control inputs $\hat{u}_{t:T}$\\
   \quad States history $\bar{\mathcal{G}} = \{G_i^\tau\}_{i=1\dots t}$\\
   \quad Time horizon $T$
   \STATE \textbf{Output: } Controls $\hat{u}_{t:T}$, predicted next time step $\hat{G}_{t+1}^\tau$ \\
   \ \\
   
   \STATE Update $A$ by descending with the gradients\\ \quad $\nabla_A\mathcal{L}_{\text{state}}(\hat{G}_t^\tau, G_t^\tau) $
   
   \STATE Forward simulation using the current graph encoding \\
       \quad$\hat{G}_{t+1}^\tau \gets \phi(G_{t}^\tau)$
       \STATE Make a buffer for storing the simulation results \\
       \quad$\mathcal{G} \gets \bar{\mathcal{G}} \cup \hat{G}_{t+1}^\tau$
       \FOR{$i=t+1, ..., T-1$}
           \STATE Forward simulation \\
           \quad $\hat{G}_{i+1}^\tau \gets \phi(\hat{G}_{i}^\tau)$; $\mathcal{G} \gets \mathcal{G} \cup \hat{G}_{i+1}^\tau$
       \ENDFOR
   	   \STATE Update $\hat{u}_{t:T}$ by descending with the gradients\\
   	   \quad $\nabla_{\hat{u}_{t:T}}\mathcal{L}_{\text{goal}}(\mathcal{G}, \mathcal{G}_g) $ \\
   \ \\
   \STATE Return $\hat{u}_{t:T}$ and $\hat{G}_{t+1}^\tau \gets \phi(G_t^\tau)$
\end{algorithmic}
\end{algorithm}
\end{center}

%% file: text/experiments.tex
\section{Experiments}

In this section, we evaluate the performance of our model on both simulation and control in three scenarios: Newton's Cradle, Rope Manipulation, and Box Pushing.
We also test how the model generalizes to new scenarios and how it learns to adapt online.

\subsection{Physics Simulation}

We aim to predict the future states of physical systems. We first describe the network used across tasks and then present the setup of each task as well as the experimental results.

\input{figText/quali.tex}
\input{figText/simu_quanti.tex}

\myparagraph{Model architecture}

For the IN baseline, we use the same network as described in the original work~\cite{Battaglia2016Interaction}. For \basemodelshort, we adopt similar network structure where the relation propagator $f_R^l(\cdot) (1\le l \le L)$ is an MLP with four 150-dim hidden layers and the object propagator $f_O^l(\cdot) (1\le l \le L-1)$ has one 100-dim hidden layer. Both output a 100-dim propagation vector. For fully observable scenarios, $f^L_O(\cdot)$ has one 100-dim hidden layer and outputs a 2-dim vector representing the velocity at the next time step. For partially observable cases, $f^L_O(\cdot)$ outputs one 100-dim vector as the latent representation.

For \modelshort, we use an MLP with three 150-dim hidden layers as the relation encoder $f_R^{\text{enc}}(\cdot)$ and one 100-dim hidden layer MLP as the object encoder $f_O^{\text{enc}}(\cdot)$. Light-weight neural networks are used for the propagators $f_O^l(\cdot)$ and $f_R^l(\cdot)$, both of which only contain one 100-dim hidden layer.

\myparagraph{Newton's cradle}
A typical Newton's cradle consists of a series of identically sized rigid balls suspended from a frame. When one ball at the end is lifted and released, it strikes the stationary balls. Forces will transmit through the stationary balls and push the last ball upward immediately. In our fully observable setup, the graph $G$ of $n$ balls has $2n$ objects representing the balls and the corresponding fixed pinpoints above the balls, as shown in~\fig{fig:cradle}a, where $n=5$. There will be $2n$ directed relations describing the rigid connections between the fixed points and the balls. Collisions between adjacent balls introduce another $2(n-1)$ relations.

We generated 2,000 rollouts over 1,000 time steps, of which 85\% of the rollouts are randomly chosen as the training set, while the rest are held as the validation set. The model was trained with a mini-batch of 32 using Adam optimizer~\cite{kingma2014adam} with an initial learning rate of 1e-3. We reduce the learning rate by 0.8 each time the validation error stops decreasing for over 20 epochs.

\fig{fig:cradle}a-c show some qualitative results, where we compare IN and \modelshort. IN cannot propagate the forces properly: the rightmost ball starts to swing up before the first collision happens. Quantitative results also show that our method significantly outperforms IN in tracking object positions. For 1,000 forward steps, IN results in an MSE of 336.46, whereas \modelshort achieves an MSE of 7.85.

\myparagraph{Rope manipulation} 

We then manipulate a particle-based rope in a 2D plane using a spring-mass model, where one end of the rope is fixed to a random point near the center and the rest of the rope is free to move. Two circular obstacles are placed at random positions near the rope and are fixed to the ground. Random forces are applied to the masses on the rope and the rope is moving in compliant with the forces. More specifically, for a rope containing $n$ particles, there will be a total of $n+2$ objects. Each pair of adjacent masses will have spring relations connecting each other, resulting in $2(n-1)$ directed edges in the dynamics graph $G$. Each mass will have a collision relation with each fixed obstacle, which adds to the graph another $4n$ edges. Frictional force applied to each mass is modeled as a directed edge connecting the mass itself.

We use the same network as described above and generate 5,000 rollouts over 100 time steps. \fig{fig:simu_quali} and \fig{fig:simu_quanti_a} show qualitative and quantitative results, respectively. We train the models with a 15-dim rope and evaluated in situations where the rope length can vary between 10 and 20. As can be seen from the figures, although the length of the underlying force propagation is fewer than Newton's Cradle's, our proposed method can still track the ground truth much more accurately and outperform IN by a large margin.

\myparagraph{Box pushing} \label{sec:exp_sim_BoxPush} In this case, we are pushing a pile of boxes forward (\fig{fig:control_quali_BoxPush}). We place a camera at the top of the scene, and only red boxes are observable. More challengingly, the observable boxes are not tracked. Therefore, the visibility of a specific box might change over time. The vertices in the graph are then defined as the state of the observable boxes and edges are defined as directional relations connecting every pair of observable boxes. Specifically, if there are $n$ observable boxes, $n(n-1)$ edges are automatically generated.
The dynamics function $\phi(\cdot)$ then takes both the scene representation and the action (\ie, position and velocity of the pusher) as input to perform an implicit forward simulation. As it is hard to explicitly evaluate a latent dynamics model,  we evaluate the downstream control tasks instead.

\myparagraph{Ablation studies}

We also provide ablation studies on how the number of propagation steps $L$ influences the final performance. Empirically, a larger $L$ can model a longer propagation path. They are however harder to train and more likely to overfit the training set, often leading to poor generalization. \fig{fig:simu_quanti_a} and~\ref{fig:simu_quanti_b} show the ablation studies regarding the choice of $L$. \modelshort achieves a high accuracy at $L=3$, with  a good trade-off between speed and accuracy. \basemodelshort achieves its best accuracy at $L=2$ but generalizes less well as $L$ increases further. This shows the benefits of using the shared encoding and residual connections used in PropNet, as described in \sect{sec:encoding_and_residual}.

\subsection{Control}

We now evaluate the applicability of the learned model on control tasks. We first describe the three tasks: Newton's Cradle, Rope Manipulation, and Box Pushing, which include both open-loop and feedback continuous control tasks, as well as fully and partially observable environments. We evaluate the performance against various baselines and test its ability on generalization and online adaptation.

\input{figText/control_quanti.tex}

\myparagraph{Newton's Cradle}

In this scenario, we assume full-state observation and a control task would be to determine the initial angle of the left-most ball, so as to let the right-most ball achieve a specific height, which can be solved with an accurate forward simulation model.

This is an open-loop control task where we only have control over the initial condition. We thus use a simplified version of Alg.~\ref{alg:control}. Given the initial physics graph and a learned dynamics model, we iteratively do forward simulation and update the control inputs by minimizing the loss function $\mathcal{L}_{\text{goal}}(\mathcal{G}, \mathcal{G}_g)$. In this specific task, the loss $\mathcal{L}_{\text{goal}}$ is the $\mathcal{L}_2$ distance between the target height of the right-most ball and the highest height that has been achieved in $\mathcal{G}$.

We initialize the swing up angle as $45^\circ$ and then optimize the angle with a learning rate of $0.1$ for $50$ iterations using Adam optimizer. We compare our model with IN. Qualitative results are shown in \fig{fig:cradle}e. Quantitatively, \modelshort's output angle has an MSE of 3.08 from the ground truth initial angle, while the MSE for interaction nets is 296.66.

\myparagraph{Rope Manipulation}

Here we define the task as to move the rope to a target configuration, where the only controls are the top two masses at the moving end of the rope (\fig{fig:control_quali}). The controller tries to match the target configuration by ``swinging'' the rope, which requires to leverage the dynamics of the rope. The loss $\mathcal{L}_{\text{goal}}$ here is the $\mathcal{L}_2$ distance between the resulting configuration and the goal configuration.

We first assume the attributes of the physics graph is known (\eg, mass, friction, damping) and compare the performance between Proportional-Derivative controller (PD)~\cite{aastrom1995pid}, Model-free Deep Reinforcement Learning (Actor-Critic method optimized with PPO~\cite{schulman2017proximal} - DRL), as well as Interaction Networks (IN) and \Model (\modelshort) with Alg.~\ref{alg:control}. \fig{fig:control_quanti} shows quantitative results, where bars marked as ``Normal'' are the results in this task (a hand-tuned PD controller has an MSE of 2.50). \modelshort outperforms the competing baselines.  \fig{fig:control_quali} shows a qualitative sample. Compared with the PD controller, our method leverages the dynamics and manages to match the target, instead of naively matching the free end of the rope.

We then consider situations where some of the attributes are unknown and can only be guessed before actually interacting with the objects. We randomly add noise of 15\% of the original scale to the attributes as the initial guesses. The ``Bias'' bars in \fig{fig:control_quanti} show that models trained with ground-truth attributes will encounter performance drop when the supplied attributes are not accurate. However, model-based methods can do online adaptation using the actual output from the environment as feedback to correct the attribute estimation. By updating the estimated attributes over the first $20$ steps of the time horizon with standard SGD, we can improve the manipulation performance so as to catch up with the situations where attributes are accurate (bars marked as ``Adapt'' in \fig{fig:control_quanti}). 

We further test whether our model generalizes to new scenarios, where the length of the rope is varied between 10 to 20. As can be seen in \fig{fig:control_quanti}, our proposed method can still achieve a good performance, even though the original \modelshort is only trained in situations with a fixed length 15 (PD has an MSE of 2.72 for generalization).

\myparagraph{Box Pushing} In this case, we aim to push a pile of boxes to a target configuration within a predefined time horizon (\fig{fig:control_quali_BoxPush}). We assume partial observation where a camera is placed at the top of the scene, and we can only observe the states of the boxes marked in red. The model trained with partial observation is compared with two baselines: DRL and IN. The loss function $\mathcal{L}_{\text{goal}}$ used for MPC is the $\mathcal{L}_2$ distance between the resulting scene encoding and the target scene encoding.

We evaluate the performance by the Chamfer Distance (CD) ~\cite{Barrow1977Parametric} between the observable boxes at the end of the episode and the target configurations, where for each box in each set, CD finds the nearest box in the other set, and sums the distance up. The negative of the distance is used as the reward for DRL. \fig{fig:control_quali_BoxPush} and \fig{fig:control_quanti_b} show qualitative and quantitative results, respectively. Our method outperforms the baselines due to its explicit modeling of the dynamics and its ability to handle multi-object interactions.

%% file: figText/quali.tex
\begin{figure*}[t]
  \centering
  \begin{subfigure}[t]{.52\linewidth}
    \includegraphics[width=\textwidth]{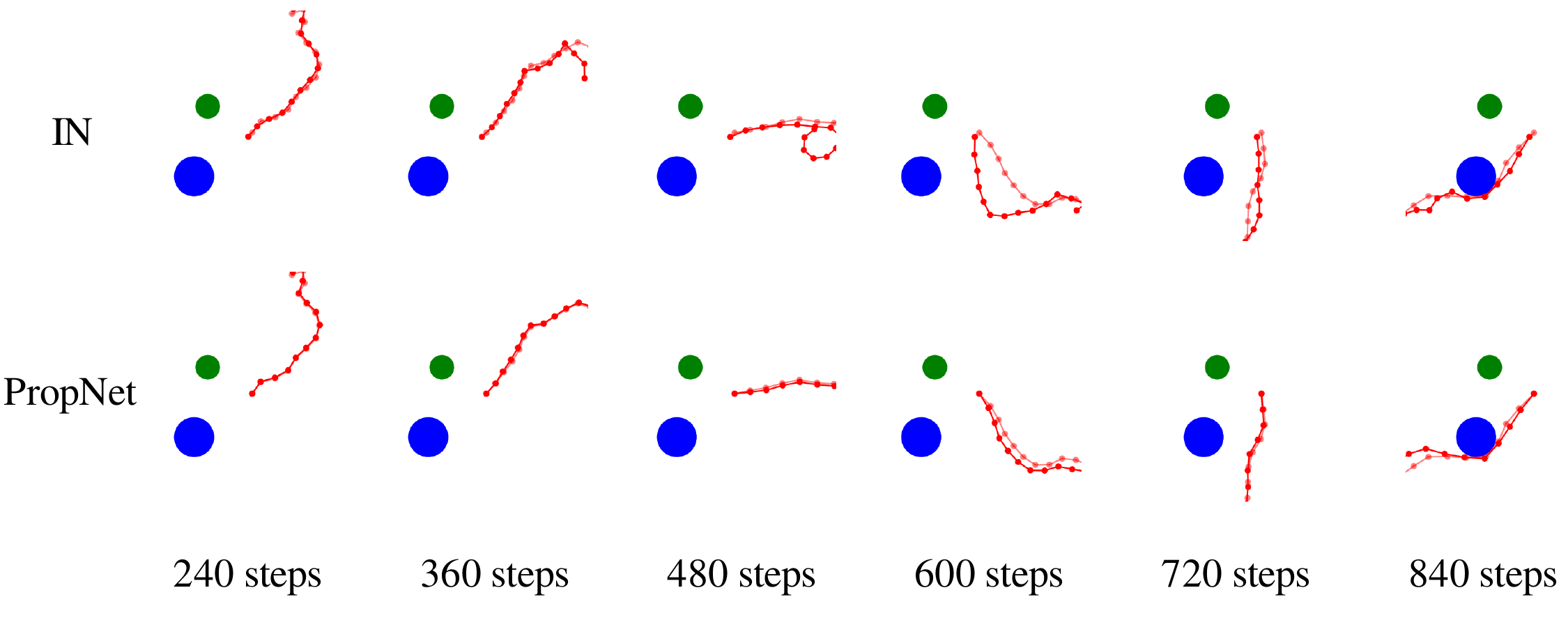}
    \caption{\label{fig:simu_quali} \footnotesize Rope Manipulation: Results on Simulation}
  \end{subfigure}
  \begin{subfigure}[t]{.47\linewidth}
    \includegraphics[width=\textwidth]{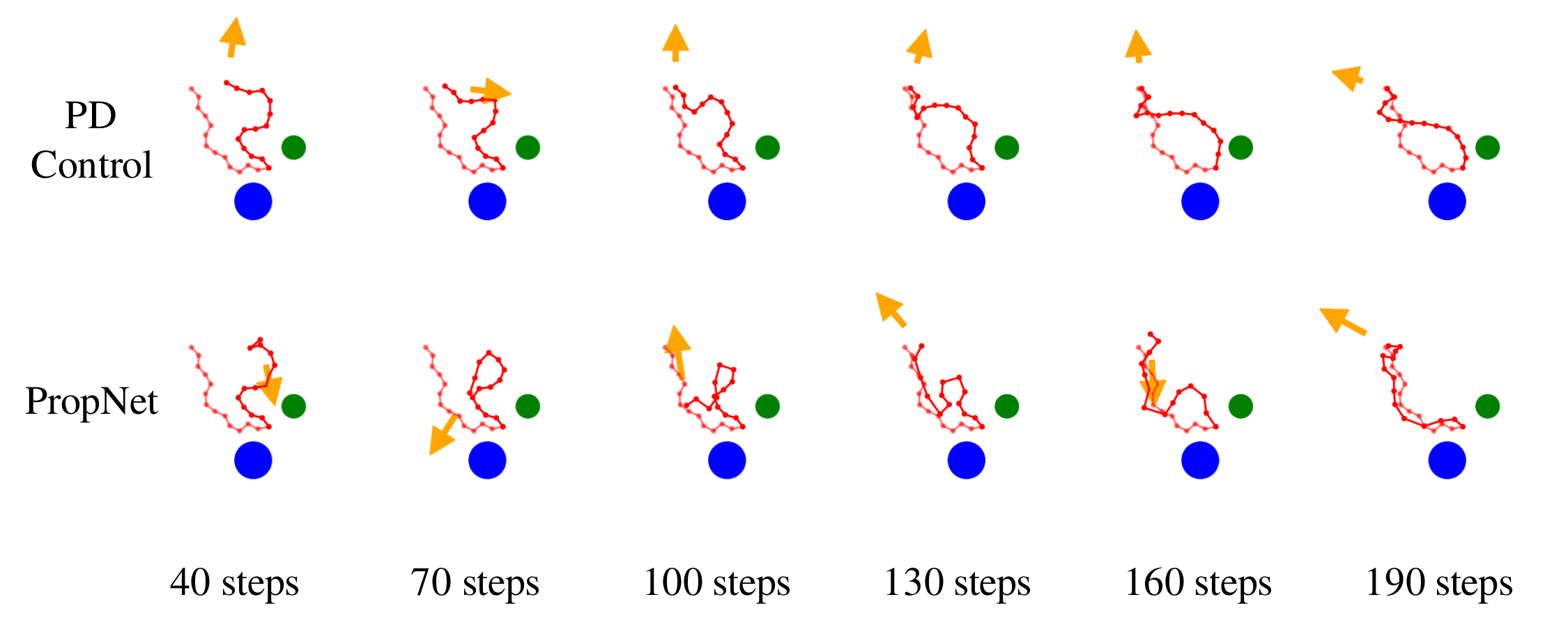}
    \caption{\label{fig:control_quali} \footnotesize Rope Manipulation: Results on Control}
  \end{subfigure}
  \begin{subfigure}{.972\linewidth}
    \centering
    \includegraphics[width=\textwidth]{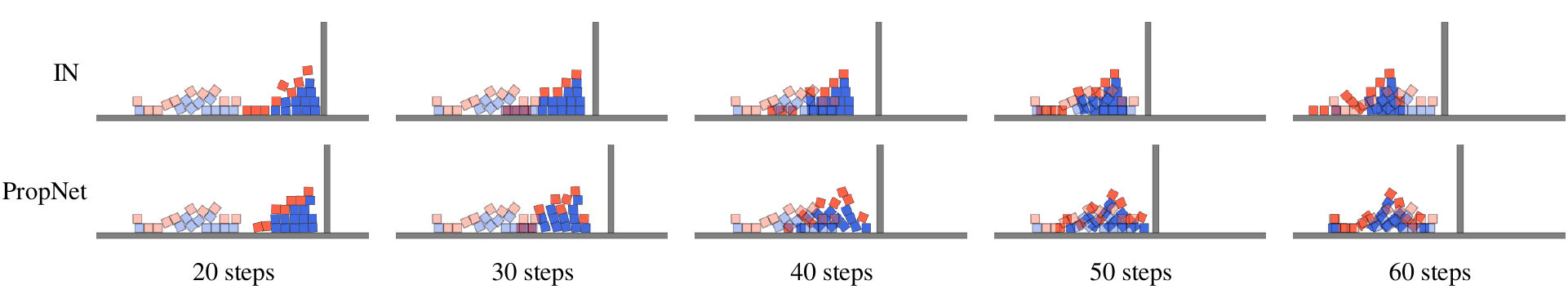}
    \caption{\label{fig:control_quali_BoxPush} \footnotesize Box Pushing: Results on Control}
  \end{subfigure}
  \caption{{\bf Qualitative results on simulation and control.} (a) Results on the planar rope simulation, where every mass on the rope has been applied a random force and the rope is moving in the planar in compliant with the forces. Our model better matches the ground truth and suffers less from the drifting problem as time horizon becomes longer. Here the transparent trajectories indicate the ground truth. (b) The rope manipulation task defines a continuous control problem which is to achieve a specified goal configuration by applying forces to the top two masses on the free end of the rope. The applied forces are visualized as yellow arrows and the goal configuration is shown as transparent. Note that instead of naively trying to match the top two masses (PD control), our control method based on \modelshort can achieve the goal configuration by exploring the rich dynamics of the rope. (c) The box pushing task requires solving a control problem under partial observation (only red blocks are observable). The goal configuration is shown as transparent. Doing control with our \model achieves more accurate outcome than with an IN. Please also see the supplementary video.}
  \label{fig:quali}
\end{figure*}

%% file: figText/simu_quanti.tex
\begin{figure}[t]
  \centering
  \begin{subfigure}[b]{.49\linewidth}
    \includegraphics[width=\textwidth]{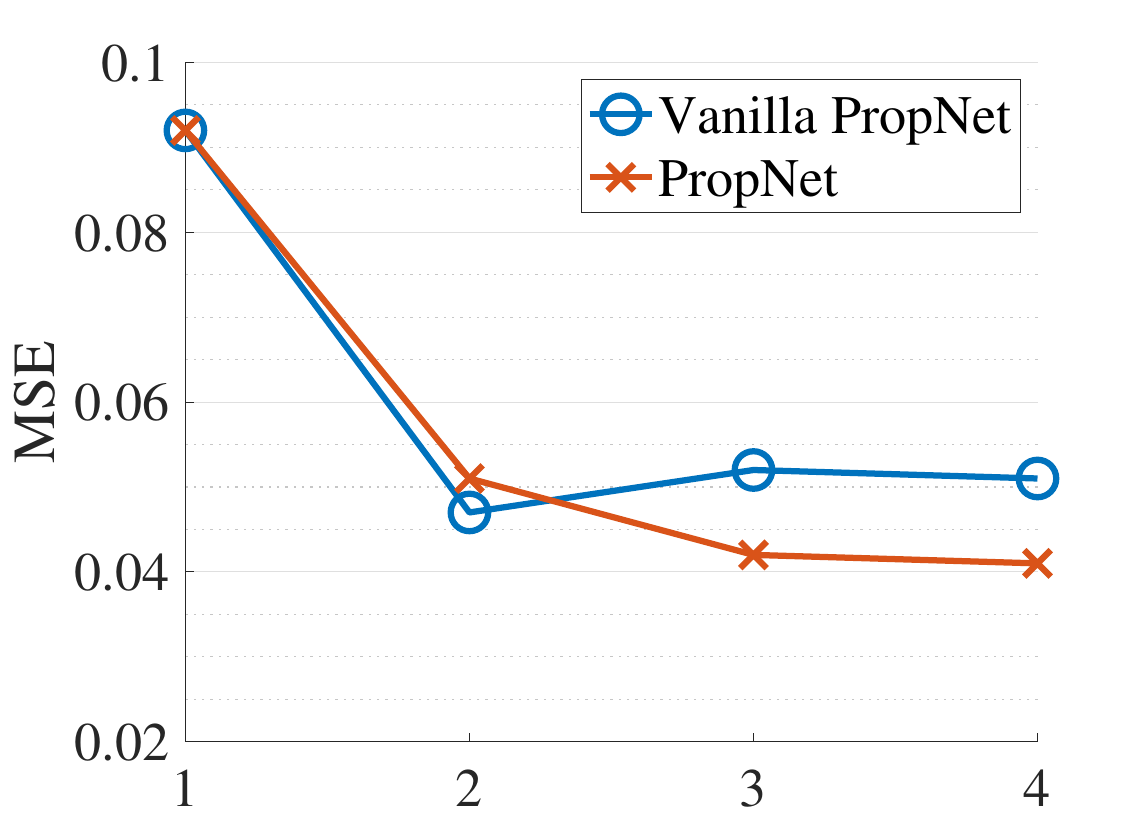}
    \caption{\label{fig:simu_quanti_a}\footnotesize Rope: Velocity}
  \end{subfigure}
  \begin{subfigure}[b]{.49\linewidth}
    \includegraphics[width=\textwidth]{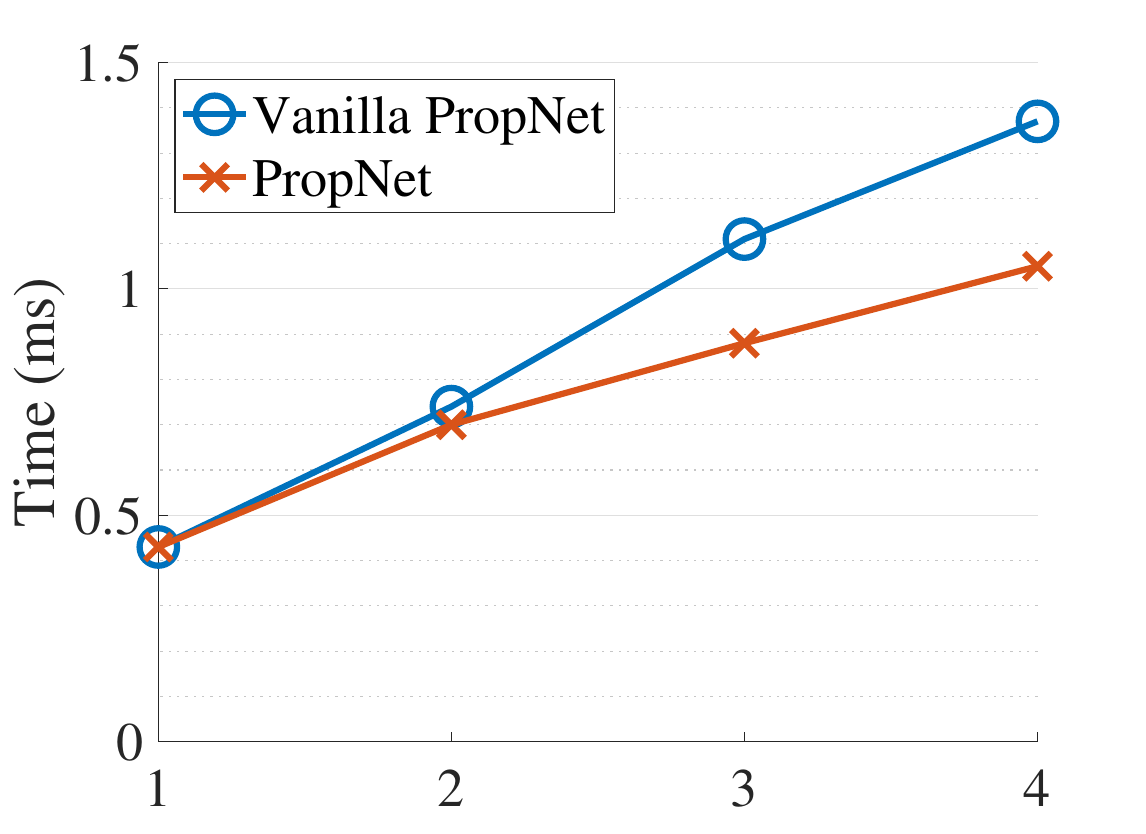}
    \caption{\label{fig:simu_quanti_b}\footnotesize Rope: Forward Time}
  \end{subfigure}
  \caption{{\bf Quantitative results on rope simulation.}
  We vary the propagation steps $L$ between $2$ to $4$ for \basemodelshort and \modelshort, which shows a trade-off between accuracy and efficiency. When $L=1$, both models reduce to Interaction Networks (IN).}
  \label{fig:simu_quanti}
\end{figure}

%% file: figText/control_quanti.tex
\begin{figure}[t]
  \centering
  \begin{subfigure}[t]{.75\linewidth}
    \includegraphics[width=\textwidth]{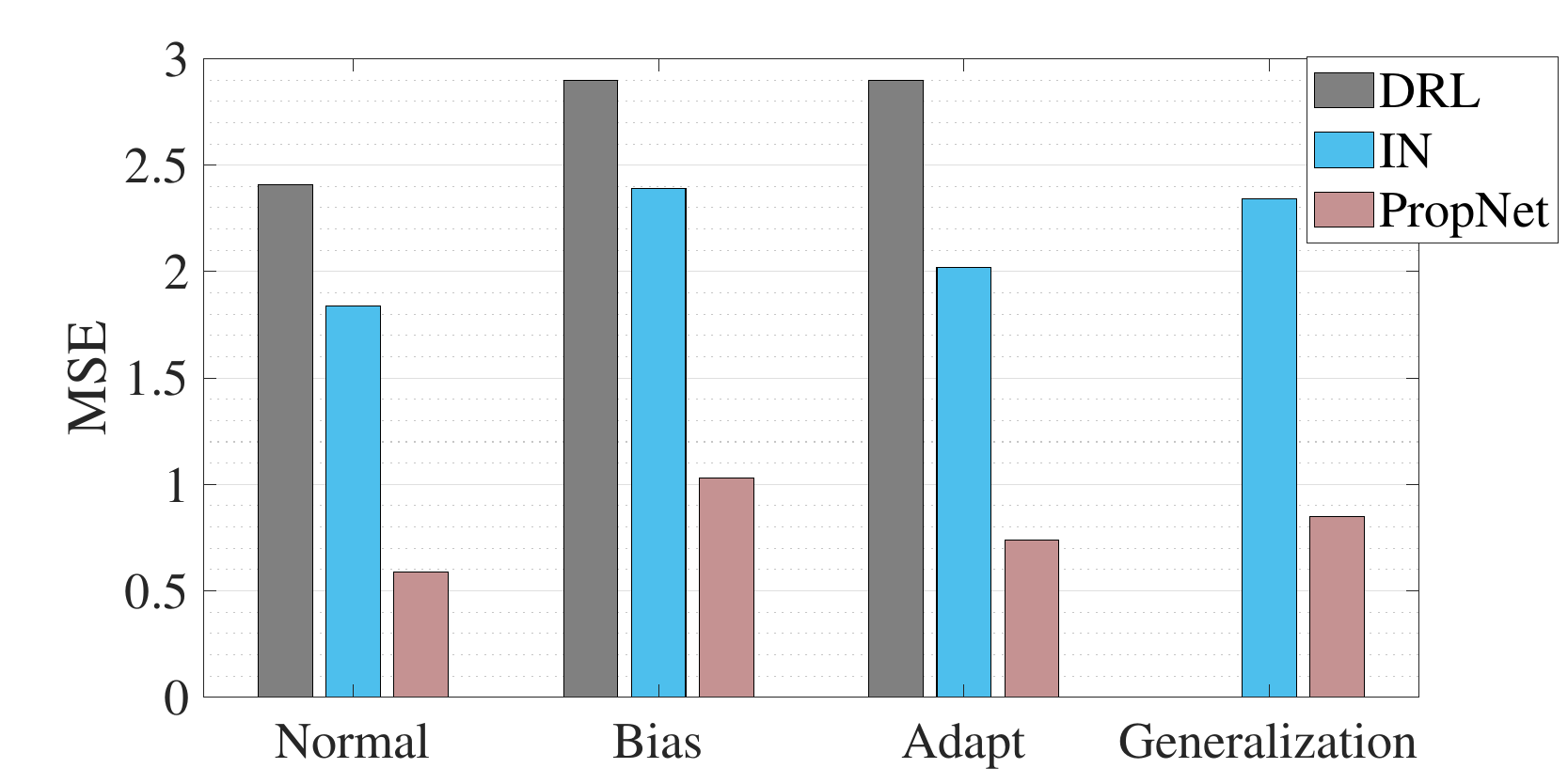}
    \caption{\label{fig:control_quanti_a} \footnotesize Rope Manipulation}
  \end{subfigure}
  \begin{subfigure}[t]{.23\linewidth}
    \includegraphics[width=\textwidth]{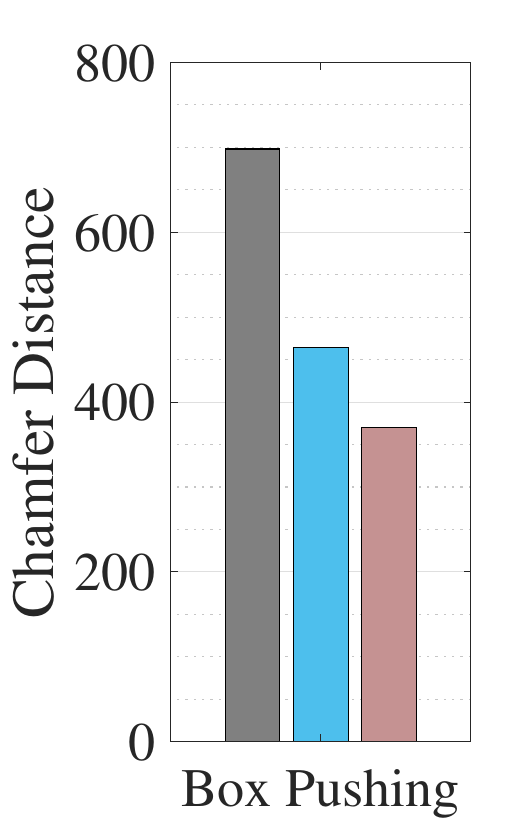}
    \caption{\label{fig:control_quanti_b} \footnotesize Box Pushing}
  \end{subfigure}
  \caption{{\bf Quantitative results on control tasks.} (a) For rope manipulation, the algorithms attempt to match a specific configuration under situations where the ground-truth attributes are known (``Normal''), where the value of the attributes are unknown (``Bias''), where algorithms actively estimate these attributes online (``Adapt''), and where ropes are of varied length between 10 to 20 when the model is only trained on ropes of length $15$ (``Genearlize''). DRL has the same performance for ``Bias'' and ``Adapt'' as it is model-free; it requires a fixed length input, and thus cannot generalize to ropes of a different length. (b) For box pushing, \model again outperforms the other methods.}
  \label{fig:control_quanti}
\end{figure}

%% file: text/conclusion.tex
\section{Conclusion}

We have presented \model (\modelshort), a general learnable physics engine that outperforms the previous state-of-the-art with a large margin. We have also demonstrated \modelshort's applicability in model-based control under both fully and partially observable environments. With propagation steps, \modelshort can propagate the effects along relations and model the dynamics of long-range interactions within a single time step. We have also proposed to improve \modelshort's efficiency by adding residual connections and shared encoding.

%% file: text/ack.tex
\section*{Acknowledgement} This work was supported by: Draper Laboratory Incorporated, Sponsor Award No. SC001-0000001002; NASA - Johnson Space Center, Sponsor Award No. NNX16AC49A; NSF \#1524817; DARPA XAI program FA8750-18-C000; ONR MURI N00014-16-1-2007; and Facebook.